\documentclass[article,10pt,twocolumn,oneside]{IEEEtran} 
\usepackage{draftwatermark}
\SetWatermarkText{2017 SoCal Robotics Symposium}
\SetWatermarkScale{0.4}
\usepackage[%
left=19.1mm,right=19.1mm,top=19.1mm,bottom=19.1mm,%
]{geometry}
\usepackage{graphicx} 
\usepackage{amsmath}
\usepackage{url}
\usepackage{caption}
\usepackage{subcaption}
\usepackage{amssymb}
\usepackage{textcomp}
\usepackage{dblfloatfix}
\usepackage{tikz}
\usetikzlibrary{spy}
\usetikzlibrary{shapes,arrows}
\tikzset{annot/.style={draw=black,fill=white,text=black}}
\title{\LARGE \bf
Optimal Needle Diameter, Shape, and Path in Autonomous Suturing$^{*}$
}

\author{Sahba Aghajani Pedram$^{1}$, Peter Ferguson$^{1}$, Ji Ma$^{1}$, Erik Dutson$^{2}$, and Jacob Rosen$^{1}$
\thanks{$^{*}$This work is supported by the U.S. National Science Foundation award IIS-1227184: Multilateral Manipulation by Human-Robot.} 
\thanks{$^{1}$Sahba Aghajani Pedram, Peter Ferguson, Ji Ma, and Jacob Rosen are with the Mechanical and Aerospace Engineering Department, University of California at Los Angeles,
	Los Angeles, CA, USA 
        {\tt\small sahbaap@ucla.edu, pwferguson@ucla.edu, jima@ucla.edu jacobrosen@ucla.edu}}%
\thanks{$^{2}$Erik Dutson is with the Department of Surgery, David Geffen School of Medicine, University of California at Los Angeles,
        Los Angeles, CA 90095, USA
        {\tt\small EDutson@mednet.ucla.edu}}%
    }
\begin{document}

\maketitle
\thispagestyle{empty}
\pagestyle{empty}

\begin{abstract}

Needle shape, diameter, and path are critical parameters that directly affect suture depth and tissue trauma in autonomous suturing. This paper presents an optimization-based approach to specify these parameters. Given clinical suturing guidelines, a kinematic model of needle-tissue interaction was developed to quantify suture parameters and constraints. The model was further used to formulate constant curvature needle path planning as a nonlinear optimization problem. The optimization results were confirmed experimentally with the Raven II surgical system. The proposed needle path planning algorithm guarantees minimal tissue trauma and complies with a wide range of suturing requirements.

\end{abstract}


\section{INTRODUCTION}
Suturing is one of the most challenging and time consuming of all surgical subtasks \cite{garcia1998manual}. 
The limitations associated with the human operator along with the repetitive nature of suturing make it a candidate for automation. The framework of automated suturing relies on the surgeon for high level decision making while delegating the execution of low-level motions to the robot, potentially improving surgical outcome.  
Many research efforts focused on generating an initial needle path which can be updated in real time to adjust to environment changes. 
In \cite{senautomating}, an algorithm was developed to minimize the suture length and maintain orthogonal needle angle at the tissue entry point. 
Some studies \cite{iyer2013single,staub2010automation} have proposed constant curvature path (CCP) planning in which the needle only rotates around its geometric center. It was argued that the CCP algorithm will result in minimal tissue trauma. However, due to the constrained motion, important suturing requirements such as suture depth may not be satisfied \cite{diao1996effect}. Needle reorientation inside the tissue is proposed in other studies \cite{jackson2013needle} to better follow suturing requirements yet it may impose additional tissue trauma \cite{nageotte2009stitching}. 
\begin{figure}[t]
	
	\centering
	
	\includegraphics[scale = 0.54, trim={7.6cm 2cm 7.7cm 4.5cm},clip]{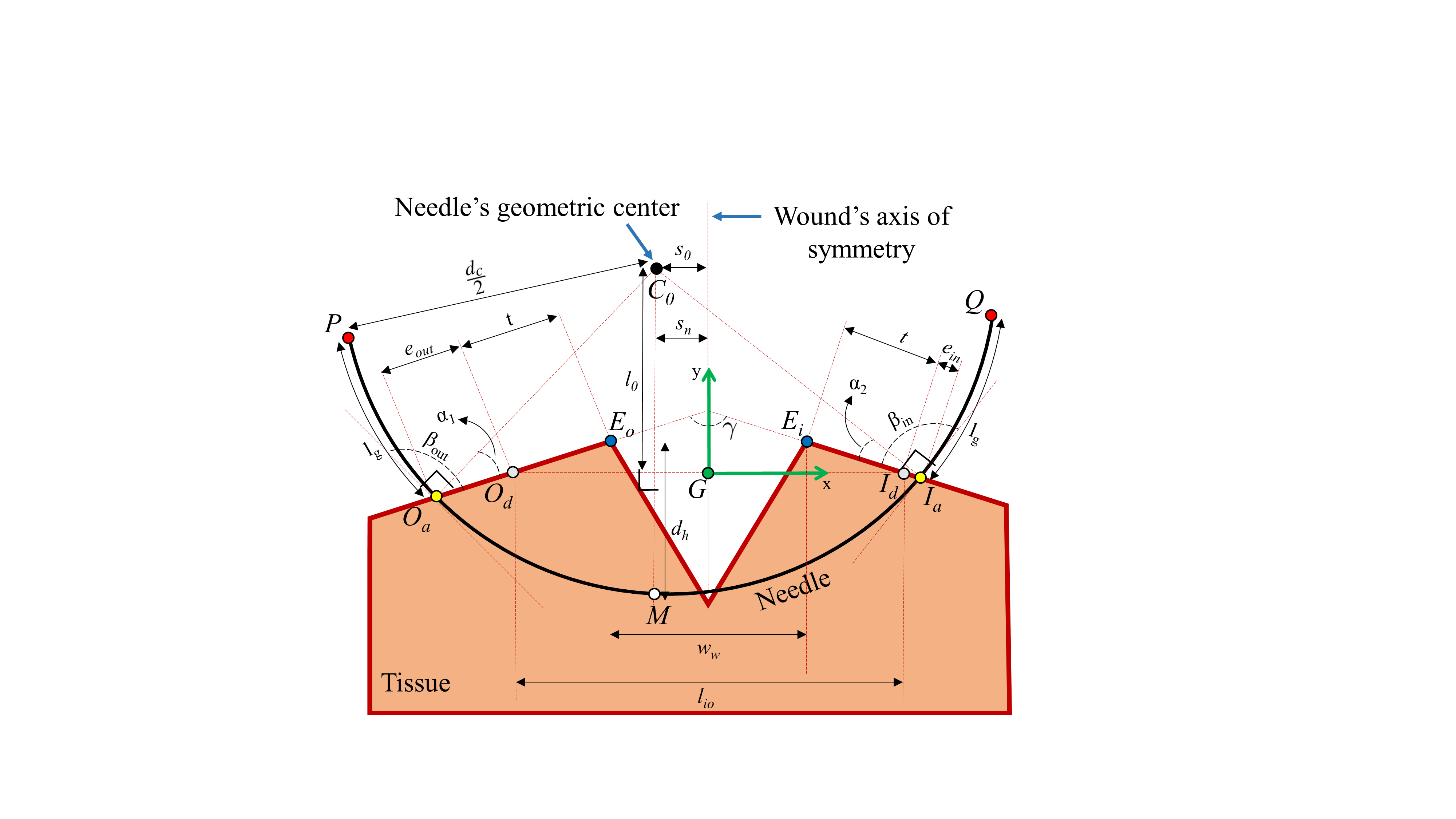}
	\caption{Kinematic model of needle-tissue interaction at switching time.}
	\label{suturegeometry}
	
\end{figure}  
Despite the fact that surgeons select needle shape and diameter based on the tissue geometry \cite{hochberg2009suture}, previous studies have largely ignored these variables in their path planning formulations. This study, which is described in greater detail in \cite{Sahba2017autonomous}, extends the current research effort by developing an algorithm that includes tissue geometry as inputs, and needle shape, diameter and path as outputs. 

\section{Method}
\subsection{Needle Motion}
This paper focuses on needle motion in which the needle rotates about its geometric center, \textit{Fixed-Center Motion (FCM)}.
There is a general consensus in the medical community that following the natural curvature of the needle produces the best suture \cite{ruurda2004manual}. 
The limitation of FCM is that any two combinations of suture parameters (defined below) will uniquely define the path. Due to this limitation, previous CCP planning algorithms only fulfilled a few suturing guidelines. We bypass this limitation by permitting, weighting, and minimizing error for all suture parameters.
\subsection{Kinematic Modeling of the Fixed-Center Motion Path}
\textbf{Assumptions:} The needle path planning can be formulated as a kinematics problem because FCM ideally creates small shear forces only. The needle is restricted to planar motion as implied by ideal FCM. The tissue geometry is assumed to be symmetric and is approximable by Fig. \ref{suturegeometry}. Lastly, access to needle shapes $\in$ \(A_n = \{\frac{1}{4},\frac{3}{8},\frac{1}{2},\frac{5}{8}\}\) is assumed. The needle shape refers to the arc length of the needle as a fraction of the circumference of a circle. 
\textbf{Suture Parameters:} Based on the suturing guidelines \cite{sherris2004essential}, we define six \textit{suture parameters} (pictured in Fig. \ref{suturegeometry}) that quantify an ideal needle path. These parameters are: 1) needle entry angle ($\beta_{in}$), 2) distance between actual and desired needle entry points ($e_{in}$), 3) needle depth ($d_{h}$), 4) needle-wound symmetry ($s_{n}$), 5) needle exit angle ($\beta_{out}$), and 6) distance between actual and desired needle exit points ($e_{out}$).  
\textbf{Needle Variables:} Each suture parameter is a function of four \textit{needle variables}: the x position of the needle center ($s_{0}$), the y position of the needle center ($l_{0}$), the needle diameter ($d_{c}$), and the needle shape ($a_{n}$). The first two needle variables are referred to as the \textit{needle position}, while the other two are referred to as the \textit{needle geometry}.   
le diameter and position uniquely identify the needle path defined by the set of all $\overrightarrow{P}$, which is given by: 
\begin{figure}[t] 	
\centering 	
\includegraphics[scale=0.4, trim={0cm 1cm 0cm 1cm},clip]{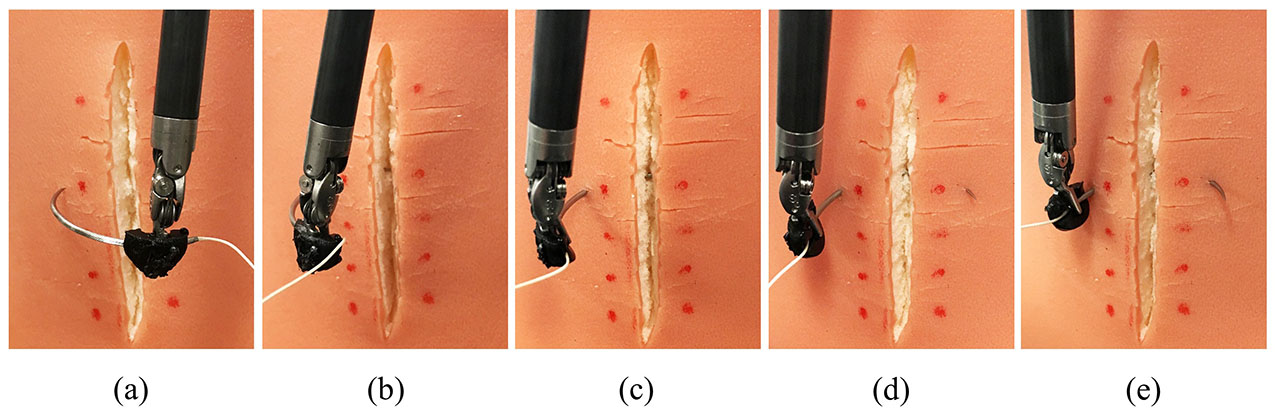}
\caption{Sequential images of automated suturing experiment.}
\label{Sequential} 	
\end{figure}
    
\textbf{Ideal Suture Parameters:} Based on the suturing guidelines, it is possible to quantify ideal values for each suture parameter. The ideal $\beta_{in}$ and $\beta_{out}$ are $\frac{\pi}{2}$. $e_{in}$, $e_{out}$, and $s_{n}$ should each be zero. $d_{h}$ is usually set to $\frac{l_{io}}{2}$, but can be changed if the surgeon prefers a different needle depth.

\textbf{Kinematic Constraints:} Kinematic constraints limit the solution set to physically viable options. For example, grasp must be able to properly switch between the two instruments, and the needle must enter the tissue from one side and exit from the other.  

 \begin{table}[b]
 	\caption{Desired, simulation, and experiment suture parameter values.}
 	\label{Error}
 	\begin{center}
 		\begin{tabular}{|c|c|c|c|}
 			\hline
 			Suture Param. & Desired & Simulation & Experiment \\
 			\hline
 			$\beta_{in}$ & 1.57 rad & 1.91 rad & 1.96 $\pm$ 0.07 rad \\
 			\hline
 			$e_{in}$ & 0 mm & 4.56 mm & 4.83 $\pm$ 0.62 mm \\
 			\hline
 			$d_{h}$ & 7.99 mm & 9.15 mm & 8.44 $\pm$ 0.73 mm \\
 			\hline
 			$s_{n}$ & 0 mm & 0 mm & 0.25 $\pm$ 0.12 mm \\
 			\hline
 			$\beta_{out}$ & 1.57 rad & 1.91 rad & 2.03 $\pm$ 0.11 rad \\
 			\hline
 			$e_{out}$ & 0 mm & 4.56 mm & 4.69 $\pm$ 0.57 mm \\
 			\hline
 		\end{tabular}
 	\end{center}
 	\label{Resutls}
 \end{table}

\textbf{Optimization Formulation:}
The CCP of the needle can be formulated as a nonlinear optimization problem of suture parameters subjected to kinematic constraints:

\begin{equation}
\begin{gathered} 
 \underset{l_{0}, s_{0}, d_{c}, a_{n}}{\text{Minimize }} J   \\ 
J = \sum\limits_{i=1}^6 \lambda_{i}(|SP_{a,i}-SP_{d,i}|) = \sum\limits_{i=1}^6 \lambda_{i} \Delta_{i} \label{18}
\end{gathered}
	\end{equation}

In \eqref{18}, $J$ is the cost function and \(\lambda_{i}\) are optimization weighting factors. \(SP_{a,i}\) and \(SP_{d,i}\) are the actual and desired values of each suture parameter, and \(\Delta_{i}\) is the absolute error between \(SP_{a,i}\) and \(SP_{d,i}\).

\textbf{Solution Algorithm:}
The equation in \eqref{18} solves for four independent needle variables. This problem is solved numerically using a brute-force search algorithm in MATLAB. 
Note that the terms ($\Delta_{i}$) in \eqref{18} have different units (rad for angles and mm for distances). Hence, each term is normalized to [0,1] to obtain the normalized cost function, $\|J\|$. 

\section{Robotic Experimental Verification:}

A Raven II surgical robot was used to evaluate the proposed algorithm. A complete sequence of the needle motion is shown in Fig. \ref{Sequential}. Similar to \cite{senautomating}, a 3D-printed jaw-mounted needle guide was utilized to enable the needle driver to securely grasp the suturing needle. The suturing is completed on a ITM-30 tissue phantom from Simulab Corp. and the phantom was manipulated to modify tissue angle.  Based on the results of the algorithm, a size 1 CTX taper point needle (Ethicon Inc.) with $a_{n}$ = $\frac{1}{2}$ and $d_{c}$ = 30.55 mm was used. The inputs of the algorithm were $\lambda_{1}$ = $\lambda_{2}$ = ... = $\lambda_{6}$ = 1, $\gamma$ = $\frac{4\pi}{5}$, $l_{io}$ = 16 mm, and $w_{w}$ = 5.5 mm. Eight trials of automated suturing were performed. The mean and standard deviation of each suture parameter is given in Table \ref{Resutls}.

\section{Discussion and Conclusion}
In this paper, we presented a constant curvature path planning algorithm which minimizes tissue trauma and optimizes needle variables to satisfy recommended suturing guidelines. The main novelty of our method is to allow and minimize errors for all the suture parameters rather than satisfying just a few, as was considered in previous studies. 
losely match the simulation results despite several sources of error. 
The accuracy of the model was tested on a tissue phantom using the Raven II. The measured results indicate the model is accurate and the assumptions are reasonable.
\bibliographystyle{IEEEtran}
\bibliography{IEEEabrv,main}
\end{document}